\documentclass[runningheads]{llncs}

\usepackage[review,year=2026,ID=xxxx]{eccv}
\usepackage{eccvabbrv}

\usepackage{graphicx}
\usepackage{booktabs}

\usepackage[accsupp]{axessibility}  % Improves PDF readability for those with disabilities.

\usepackage{hyperref}

\usepackage{orcidlink}

\usepackage{xcolor}
\usepackage{pifont}
\usepackage{stfloats}
\usepackage{wrapfig}
\usepackage{graphicx}
\usepackage{algorithm}        % 提供浮动体 Algorithm
\usepackage{algorithmicx}     % 提供 algorithmicx 框架
\usepackage{algpseudocode}

\begin{document}

\title{Tele-Catch: Adaptive Teleoperation for Dexterous Dynamic 3D Object Catching} 

\titlerunning{Abbreviated paper title}

\author{First Author\inst{1}\orcidlink{0000-1111-2222-3333} \and
Second Author\inst{2,3}\orcidlink{1111-2222-3333-4444} \and
Third Author\inst{3}\orcidlink{2222--3333-4444-5555}}

\authorrunning{F.~Author et al.}

\institute{Princeton University, Princeton NJ 08544, USA \and
Springer Heidelberg, Tiergartenstr.~17, 69121 Heidelberg, Germany
\email{lncs@springer.com}\\
\url{http://www.springer.com/gp/computer-science/lncs} \and
ABC Institute, Rupert-Karls-University Heidelberg, Heidelberg, Germany\\
\email{\{abc,lncs\}@uni-heidelberg.de}}

\maketitle

\vspace{-1.5em}

\begingroup
\renewcommand\thefootnote{$\dagger$}
\footnotetext{\small{Corresponding author}}
\endgroup
\begin{abstract}
Teleoperation is a key paradigm for transferring human dexterity to robots, yet most prior work targets objects that are initially static, such as grasping or manipulation. Dynamic object catch, where objects move before contact, remains underexplored. Pure teleoperation in this task often fails due to timing, pose, and force errors, highlighting the need for shared autonomy that combines human input with autonomous policies. To this end, we present Tele-Catch, a systematic framework for dexterous hand teleoperation in dynamic object catching. At its core, we design DAIM, a dynamics-aware adaptive integration mechanism that realizes shared autonomy by fusing glove-based teleoperation signals into the diffusion policy denoising process. It adaptively modulates control based on the interaction object state. To improve policy robustness, we introduce DP-U3R, which integrates unsupervised geometric representations from point cloud observations into diffusion policy learning, enabling geometry-aware decision making. Extensive experiments demonstrate that Tele-Catch significantly improves accuracy and robustness in dynamic catching tasks, while also exhibiting consistent gains across distinct dexterous hand embodiments and previously unseen object categories.
\end{abstract}

\section{Introduction}
\label{sec:intro}
Robotic teleoperation is a fundamental paradigm for transferring human dexterity to robotic systems~\cite{honerkamp2025whole, he2024learning,fu2024mobile,hagita2024cybernetic}, enabling applications in human–robot collaboration, remote operation, and industrial automation. Most existing research has concentrated on manipulating objects that are initially static, including grasping, manipulation, and in-hand adjustment~\cite{huang2025dih,lidexdeform, yin2025dexteritygen}. In contrast,  dynamic object interaction, in particular the real-time catching of free-falling or moving objects, remains underexplored despite its importance, as such capability is frequently required in human–object interaction and manipulation tasks. Furthermore, whereas fully automated policies are typically limited to predefined behavior patterns, shared autonomy with teleoperation offers the flexibility to adapt to diverse tasks based on human intent. In this work, we conduct systematic studies of multi-fingered dexterous hand teleoperation for dynamic object catching.

\begin{figure}[t] 
  \centering
  \includegraphics[width=0.94\textwidth]{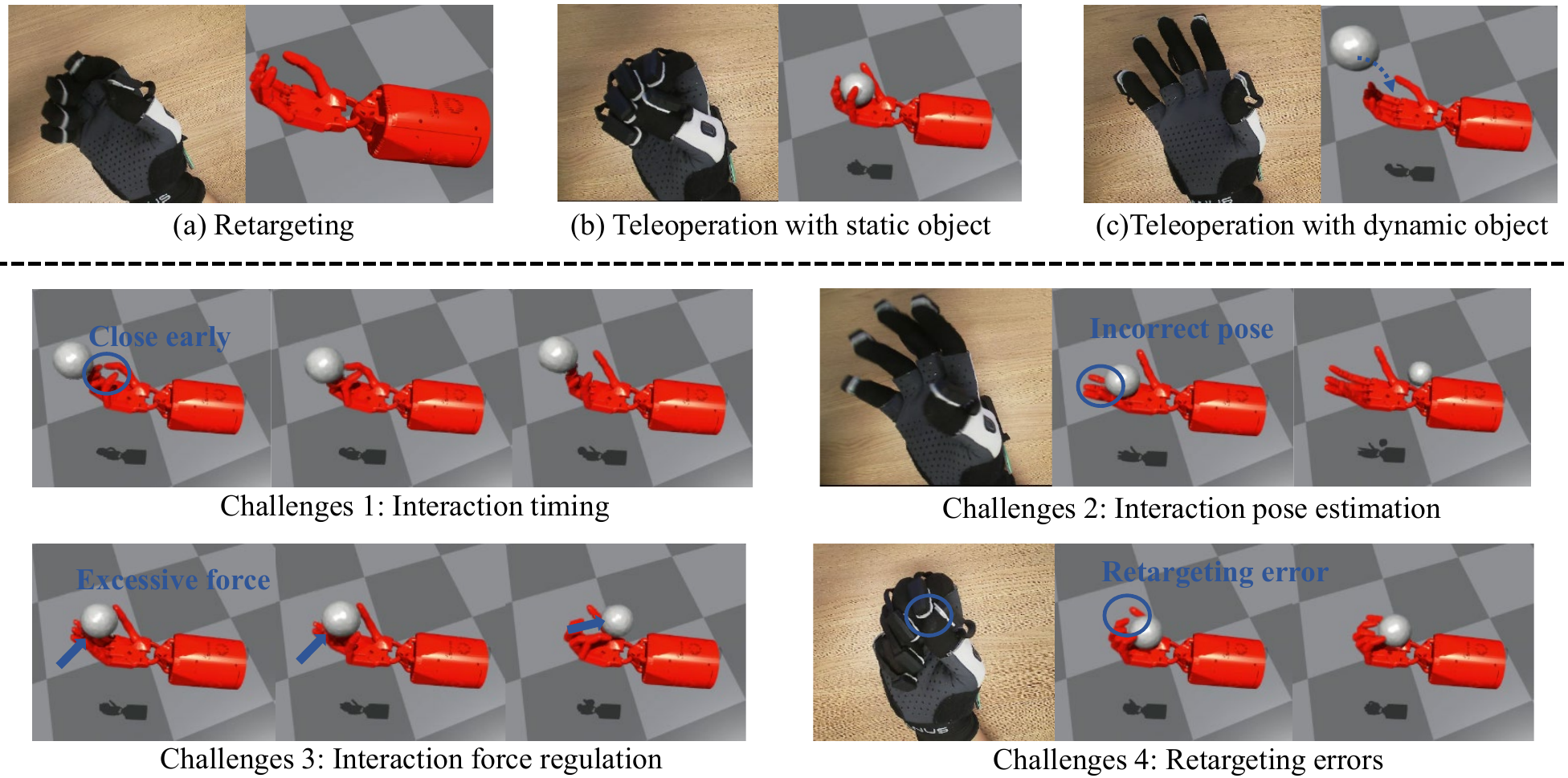} 
  \caption{Teleoperation in Dexterous Manipulation. (a) Retargeting maps human hand motions to robotic hands for teleoperation. (b) Teleoperation with static object interaction. (c) Teleoperation with dynamic object interaction (our focus). Four representative cases illustrate the key challenges in teleoperation with dynamic objects: (1) grasping too early, (2) incorrect grasp pose, (3) excessive grasping force, and (4) retargeting errors between human and robotic hands, all of which eventually cause the object to slip or fall.} 
  \vspace{-3.0em}
  \label{fig:banner}
\end{figure}

Despite its promise, pure teleoperation (based solely on retargeting) for dynamic object catching faces several fundamental challenges, as summarized in Fig.~\ref{fig:banner}. (1) Interaction timing is highly sensitive in dynamic scenarios, where even slight delays or premature actions frequently lead to failure. (2) Interaction pose estimation remains challenging, as rapid object movement hinder the reliable perception of feasible contact configurations. (3) Interaction force regulation is difficult, since insufficient force often results in slippage, while excessive force destabilizes the manipulation. (4) Retargeting errors arise from the structural mismatch between human hands and robotic hands, which become more pronounced under fast motions~\cite{yin2025geometric}. To this end, we turn to learning-based frameworks as a potential foundation for reliable control. Reinforcement Learning (RL)~\cite{landexcatch} and Diffusion Policy (DP)~\cite{chi2023diffusion,Ze2024DP3} demonstrate strong capabilities in acquiring stable dynamic manipulation strategies, while teleoperation provides adaptability and flexibility. This motivates the inquiry into whether teleoperation inputs can be incorporated into learned frameworks to ensure robustness and stability in dynamic object catching.

In this regard, we introduce Tele-Catch, a shared-autonomy framework for dexterous hand teleoperation in dynamic object catching. The core of our shared autonomy is DAIM, a \textbf{d}ynamics-aware \textbf{a}daptive \textbf{i}ntegration \textbf{m}echanism that fuses glove-based teleoperation inputs into the diffusion denoising steps while adapting the integration strength to the object’s dynamic state. In particular, the cosine schedule used in DAIM allows the teleoperation influence to increase smoothly over diffusion steps, avoiding abrupt changes and ensuring stable shared autonomy. This design yields a smooth and flexible control scheme that balances robustness from the learned policy with adaptability from teleoperation input. 

To further enhance policy robustness and generalization, we develop DP-U3R, which augments the diffusion policy with unsupervised 3D point-cloud representations. Specifically, point-cloud observations are perturbed with Gaussian noise and processed by an encoder to extract both point-wise geometric features and pooled global structure, which are combined through attention to yield geometry-aware representations. To save the calculation cost, only global structure features are integrated into the observation space of the diffusion policy, enabling it to leverage geometric priors for more accurate action generation. Extensive experiments on mainstream data and robotic platforms validate the effectiveness and generalization of our method, showing substantial improvements in teleoperated dynamic object catching. The contributions of our work are as follows:

\begin{itemize}
\item We introduce Tele-Catch, pioneering the systematic study of dexterous hand teleoperation for dynamic object catching and opening a new direction beyond static grasping tasks.

\item We design a dynamics-aware adaptive integration mechanism that smoothly blends glove teleoperation with learned policies, improving stability under diverse motion conditions.

\item We develop a concise yet effective modular DP-U3R, which incorporates unsupervised point-cloud geometric representations into the diffusion policy, enhancing its robustness and generalization.

\item Experiments conducted on representative data and hardware platforms confirm the robustness and transferability of our approach, yielding notable gains in teleoperation.
\end{itemize}

\section{Related Work}

Tab.~\ref{tab:related} provides an overview of representative methods that are closest to our task and method. Prior work either tackles dynamic catching with low-DoF end-effectors or focuses on static dexterous manipulation with limited generalization and without shared autonomy. In contrast, our approach uniquely combines adaptive shared autonomy, 3D perception, and multi-hand generalization for dynamic dexterous catching. We elaborate on the detailed differences in the remainder of this section.

\begin{table}[t]
\centering
\vspace{-2mm}

\newcommand{\cmark}{\textcolor{green!60!black}{\checkmark}}
\newcommand{\xmark}{\textcolor{red!70!black}{\ding{55}}}

\resizebox{\textwidth}{!}{
\begin{tabular}{lccccccccc}
\toprule
Method & Venue & Shared-Aut. & Deployment & Dynamic-Cat. & Visual & Unseen-Obj. & Hand-Types & DOF  \\
\midrule
DSA~\cite{salehian2016dynamical} & TRO'16  
& \xmark & \textcolor{green!60!black}{Sim.+Real} & \cmark & \textcolor{green!60!black}{2D} 
& \xmark & \textcolor{red!70!black}{Single}
& \textcolor{red!70!black}{16}
 \\

Bi-DexHands~\cite{chen2023bi} & TPAMI'23 
& \xmark  & \textcolor{red!70!black}{Sim.} & \cmark & \textcolor{green!60!black}{3D} 
& \xmark & \textcolor{red!70!black}{Single}
& \textcolor{red!70!black}{16}
\\

DP3~\cite{Ze2024DP3} & RSS'24
& \xmark  & \textcolor{green!60!black}{Sim.+Real}  & \xmark & \textcolor{green!60!black}{3D} 
& \xmark & \textcolor{red!70!black}{Single}
& \textcolor{red!70!black}{16}
 \\

DexterityGen~\cite{yin2025dexteritygen} & RSS'25
& \textcolor{red!70!black}{Fixed}  & \textcolor{green!60!black}{Sim.+Real}  & \xmark & \xmark
& \cmark & \textcolor{red!70!black}{Single}
& \textcolor{red!70!black}{16}
 \\

DexHandDiff~\cite{liang2025dexhanddiff} & CVPR'25
& \xmark  & \textcolor{red!70!black}{Sim.} & \xmark & \xmark
& \xmark & \textcolor{red!70!black}{Single}
& \textcolor{green!60!black}{24}
 \\

\textbf{Tele-Catch (Ours)} & --
& \textcolor{green!60!black}{Adaptive}  & \textcolor{green!60!black}{Sim.+Real} & \cmark & \textcolor{green!60!black}{3D}
& \cmark & \textcolor{green!60!black}{Multi}
& \textcolor{green!60!black}{16/24}
 \\
\bottomrule
\end{tabular}
}
\caption{Comparison with main related work. Shared-Aut. is Shared-Autonomy. And Obj. stands for Objects.  Dynamic-Cat. denotes  Dynamic-Catching Task. Sim. and Real denote deployment in simulation and on real robots, respectively.}
\vspace{-3em}
\label{tab:related}
\end{table}

\textbf{Teleoperation and Shared Autonomy.} Teleoperation is an important paradigm for transferring human dexterity to robotic systems, and existing approaches include motion-based~\cite{xia2024teleoperation, zhang2025doglove}, vision-based~\cite{handa2020dexpilot,li2022dexterous,cheng2025open}, and exoskeleton-based~\cite{chao2025exo, dong2025gex} methods. For instance, Doglove~\cite{zhang2025doglove} enables low-cost dexterous hand control without additional visual systems or learned policies. ByteDexter~\cite{wen2025dexterous} uses the Manus glove with optimization-based retargeting to control a 20-DoF hand for real-time in-hand manipulation and long-horizon tasks. Furthermore, DexterityGen~\cite{yin2025dexteritygen} injects teleoperation weights into diffusion denoising for in-hand manipulation, but its fixed hard threshold can cause abrupt arbitration, and it does not explicitly leverage visual observations to ground denoising.

 In contrast, our Tele-Catch introduces a dynamics-aware adaptive integration mechanism that enables smooth and stable coordination between glove-based teleoperation and learned policies (Eq.~11--14), where the intervention is scheduled along denoising steps and further adjusted by object dynamics. We additionally incorporate 3D point-cloud observations (via DP-U3R) to provide geometry-aware guidance, improving stability under varying motion conditions. On the other hand, most prior work~\cite{wen2025dexterous,yin2025dexteritygen} still focuses on static object manipulation, and very few studies explore teleoperation with dynamic objects such as catching or handover. To fill this gap, we introduce Tele-Catch, a systematic study of dexterous hand teleoperation for dynamic object catching.

\textbf{Dexterous Hand Manipulation.} Recent research on dexterous hand manipulation can be categorized into control-driven, learning-driven, and human-in-the-loop. Control-driven approaches~\cite{salehian2016dynamical, jiang2024contact, hess2024sampling} rely on explicit physical models and advanced controllers such as model predictive control to manage multi-contact dynamics, enabling stable in-hand manipulation in both simulation and real-world tests. Learning-driven approaches~\cite{yuancross, li2025maniptrans} leverage data-driven strategies, including RL, imitation learning, and diffusion-based generative models, achieving robust dexterity and improved generalization across diverse manipulation tasks. Finally, human-in-the-loop approaches~\cite{wangsparsedff,liudextrack} integrate human guidance through teleoperation interfaces—such as motion-capture gloves, vision-based tracking, or exoskeleton devices—providing intuitive control of dexterous hands and generating high-quality demonstrations that enhance adaptation for complex manipulation scenarios. In this work,  we attempt to optimize the learning policy to assist and enhance teleoperation.

\textbf{Point Cloud Representation for Embodied AI.} Recent works~\cite{chen2025clutterdexgrasp,yang2025fp3} recognize that 3D point clouds, with their inherent state representations, effectively improve multi-fingered hand manipulation and teleoperation. Specifically, DP3~\cite{Ze2024DP3} develops a visual imitation learning algorithm that integrates 3D visual representations with diffusion policies. It effectively handles various robotic tasks and achieves strong generalization with few demonstrations. In addition, ViViDex~\cite{chen2025vividex} proposes a vision-based framework that transforms depth-based point clouds into hand-centric coordinates to better capture hand–object geometry, enabling robust visual policies for multi-finger control without relying on privileged object states. In human-in-the-loop settings, teleoperation systems have begun to incorporate point clouds for state estimation and policy learning – for example, CordViP~\cite{fu2025cordvip} utilizes accurate object pose and robot proprioception to produce interaction-aware point clouds of the hand and object, and pretrains an observation encoder on hand–object contact correspondences to guide a DP. In contrast, our work seeks to obtain self-supervised geometric features from dynamic object point clouds and incorporate them into the DP to enhance the system robustness.

\section{Preliminaries}
\textbf{Reinforcement Learning (RL).} RL provides a framework for sequential decision-making, where an agent learns to maximize long-term returns through interaction with the environment~\cite{chen2022towards}. At each timestep $t$, the agent observes a state $s_t$, samples an action $a_t \sim \pi_\theta(a_t|s_t)$, receives a reward $r_t$, and transitions to the next state $s_{t+1}$. The optimization objective is the expected discounted return$J(\theta) = \mathbb{E}_{\pi_\theta}\left[\sum_{t=0}^T \gamma^t r_t \right]$, where $\gamma \in (0,1)$ denotes the discount factor. Proximal Policy Optimization (PPO)~\cite{schulman2017proximal} is a widely used algorithm for continuous control, improves upon standard policy gradients with a clipped surrogate objective that stabilizes training by constraining policy updates. PPO updates the RL policy by the clipped surrogate loss:

{\small
\begin{equation*}
L^{\text{CLIP}}(\theta) = \mathbb{E}_t \left[ 
    \min \!\big( r_t(\theta)\hat{A}_t,\; 
    \text{clip}(r_t(\theta), 1-\epsilon, 1+\epsilon)\hat{A}_t \big) 
\right],
\end{equation*}
}
with probability ratio $r_t(\theta) = \frac{\pi_\theta(a_t|s_t)}{\pi_{\theta_\text{old}}(a_t|s_t)}$ and advantage estimate $\hat{A}_t$. The reward function encodes task-specific objectives, thereby shaping the learned policy behavior.

\textbf{Diffusion Policy (DP).} DP~\cite{chi2023diffusion} is built upon the framework of the denoising diffusion probabilistic model (DDPM)~\cite{ho2020denoising}, where the diffusion mechanism is applied to action trajectories instead of raw data. DDPM processes data by gradually adding Gaussian noise and training a neural network to reverse this process. In inference, a sample is drawn from pure noise and iteratively denoised. DP adapts this principle to action trajectories, enabling multimodal and temporally coherent control generation. Formally, given a clean sample $x_0$, the forward diffusion process produces noisy versions $x_k$:
{\small
\begin{equation}
q(x_k \mid x_{k-1}) = \mathcal{N}\!\left(\sqrt{1-\beta_k}\,x_{k-1},\, \beta_k I\right),
\end{equation}
}
where $\beta_k$ is the variance schedule at step $k$. After sufficient steps ($k = K$), the data becomes nearly Gaussian noise. The reverse process, parameterized by a neural network $\epsilon_\theta$, predicts either the clean data or the added noise:
{\small
\begin{equation}
p_\theta(x_{k-1} \mid x_k) = \mathcal{N}\!\left(\mu_\theta(x_k, k),\, \Sigma_\theta\right).
\end{equation}
}
By iteratively denoising from noise $x_K \sim \mathcal{N}(0,I)$, DDPMs recover samples consistent with the training distribution. This mechanism provides the foundation for DP.

\begin{figure*}[ht] 
  \centering
  \vspace{-0.5em}
  \includegraphics[width=0.9775\textwidth]{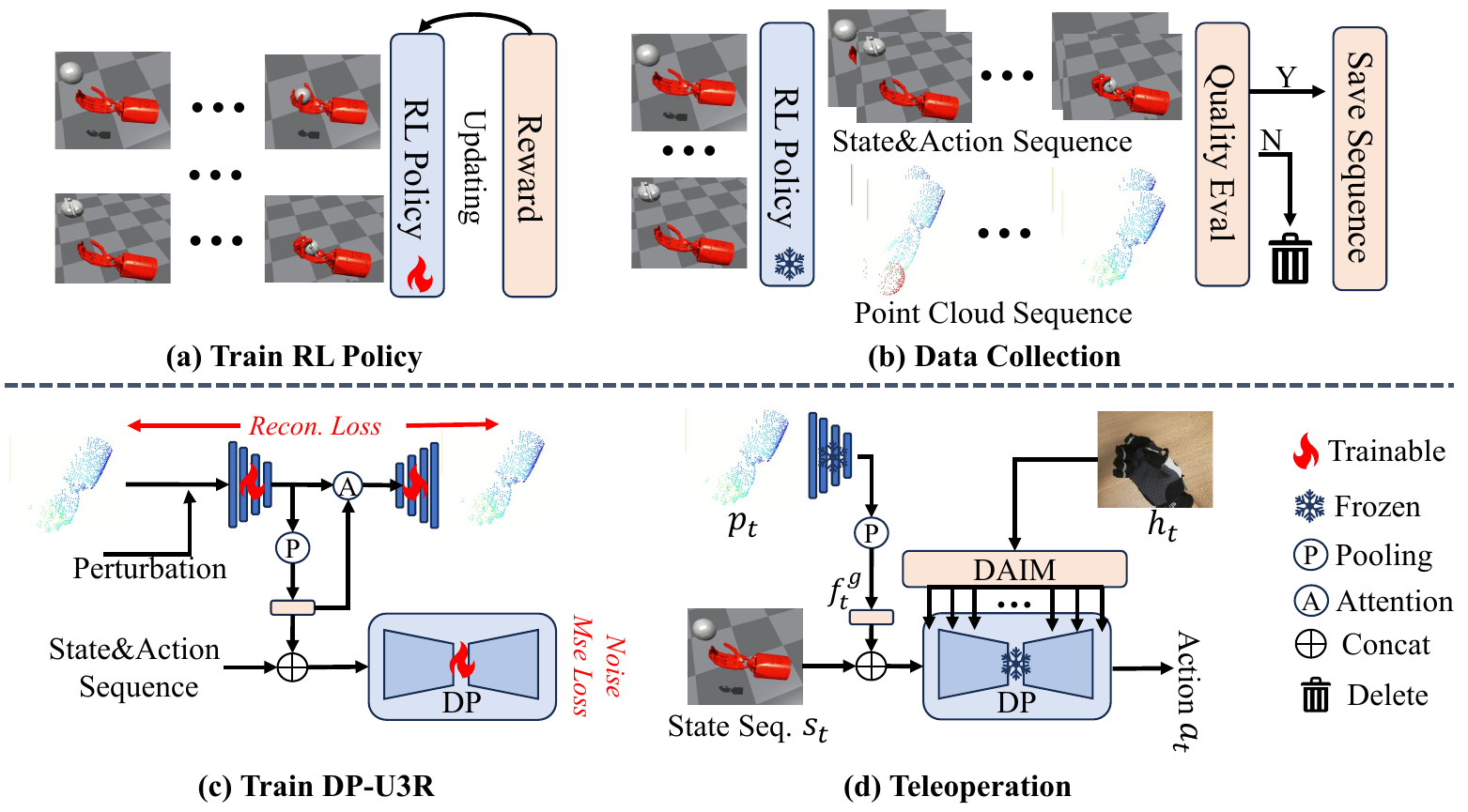} 
  \vspace{-0.5em}
  \caption{Our Method. For clarity, we present our framework by dividing it into four components: (a)~Train RL Policy (b)~Data Collection (c)~Train DP-U3R (d)~Teleoperation. Recon. stands for Reconstruction and Seq. is Sequence.} 
  \vspace{-1.5em}
  \label{fig:net}
\end{figure*}

\section{Method}
\subsection{Problem statement}
We investigate dynamic object catching under teleoperation, where a dexterous robotic hand, controlled by a real-world teleoperation glove, is required to capture moving objects within a simulated environment. Let the state and point cloud at time $t$ be $s_t \in \mathcal{S}$ and $p_t \in \mathcal{P}$, the glove input $h_t \in \mathcal{H}$, and the hand action $a_t \in \mathcal{A}$, with transitions governed by $s_{t+1} \sim P(s_{t+1} \mid s_t,a_t)$, which encode objectives such as successful catching and stability. The objective of Tele-Catch is to learn a shared-autonomy policy $\pi_\theta(a_t \mid s_t, p_t, h_t)$ that enables a dexterous hand to stably catch free-falling or dynamically moving objects. Given the environment state $s_t$, point cloud $p_t$, glove input $h_t$, and hand action $a_t$, the policy is trained to ensure robustness and adaptability under dynamic conditions.

\subsection{Overview}
As shown in Fig.~\ref{fig:net}, our framework consists of four stages. (a) We first train an RL policy in simulation to acquire stable dynamic catching skills. (b) The trained policy is then adopted to collect large-scale trajectories of state–action pairs and point cloud observations, which are filtered by the designed quality evaluation. (c) These collected data are used to train our DP-U3R, where point clouds are perturbed, encoded, and fused via pooling and attention to provide geometry-aware features that enhance the diffusion policy. (d) During teleoperation, we develop the DAIM that injects glove signals into the denoising process, adaptively balancing DP-derived policy stability with human guidance based on object dynamics and diffusion steps.

\subsection{Train RL Policy}
In the first stage, we employ the PPO algorithm to train an RL policy for dynamic object catching. PPO is selected due to its stability and effectiveness in high-dimensional continuous control, making it well-suited for dexterous hand manipulation tasks. To guide learning, we design a composite reward function tailored to the catching task, which integrates distance, orientation, and fingertip--object contact terms, along with penalties on fingertip velocities, action magnitude, torque, and energy consumption. A one-time drop penalty is further introduced for catastrophic failure. The complete reward is defined as a weighted sum:  
{\small
\begin{equation}
\begin{aligned}
R &= \lambda_{\text{dist}} \, R_{\text{dist}} 
   + \lambda_{\text{rot}} \, R_{\text{rot}} 
   + \lambda_{\text{ftip-dist}} \, R_{\text{ftip-dist}}\\
  &\quad  - \lambda_{\text{drop}} \, P_{\text{drop}} - \lambda_{\text{lin}} \, P_{\text{ftip-linvel}} 
        - \lambda_{\text{ang}} \, P_{\text{ftip-angvel}} \\
  &\quad - \lambda_{\text{act}} \, P_{\text{action}}
        - \lambda_{\text{torque}} \, P_{\text{torque}}
        - \lambda_{\text{power}} \, P_{\text{power}} \, ,
\end{aligned}
\label{eq:reward}
\end{equation}
}
where $R_{\text{dist}}, R_{\text{rot}}, R_{\text{ftip-dist}}$ represent distance, orientation alignment, and fingertip--object distance rewards, respectively. The penalty terms $P_{\text{ftip-linvel}}, P_{\text{ftip-angvel}}$, $P_{\text{action}}$, $P_{\text{torque}}$, $P_{\text{power}}$ regulate motion smoothness, torque load, and energy efficiency, while $P_{\text{drop}}$ penalizes object fall. Each component is weighted by its coefficient $\lambda$, which specifies its relative importance. The exact numerical values of all coefficients are summarized in Appendix.~\ref{sec:reward}. 

\subsection{Data Collection}
We construct a dataset of dynamic catching trajectories using the trained RL policy. 
Let the environment state at time $t$ be $s_t \in \mathcal{S}$, which contains the object pose, velocity, and proprioceptive information of the robotic hand. A depth camera is mounted in the environment to provide a point cloud observation $p_t \in \mathcal{P}$, representing the 3D geometry of the scene at time $t$. The hand action is denoted by $a_t \in \mathcal{A}$, corresponding to the joint commands executed by the robotic hand. 
Thus, each trajectory can be formalized as a sequence 
{\small
\begin{equation}
\tau = \{(s_t, p_t, a_t)\}_{t=0}^T .
\end{equation}
}
To ensure the quality of the collected data, we define a quality evaluation function $Q(\tau)$ for each trajectory. 
A trajectory is accepted only if it satisfies two conditions: 
(i) the object is not dropped during the interaction, and 
(ii) the final object--palm distance is below a predefined threshold $\delta$. 
Formally,
{\small
\begin{equation}
Q(\tau) = \mathbf{1} \Big[ \text{no\_drop}(\tau) \ \land \ \| p_{\text{T}}^{obj} - p_{\text{T}}^{palm} \|_2 < \delta \Big] ,
\end{equation}
}
where $p_{\text{T}}^{obj}$ and $p_{\text{T}}^{palm}$ denote the positions of the object and the palm center at the final timestep $T$. 
Only trajectories with $Q(\tau) = 1$ are retained and stored, while the rest are discarded. This quality-controlled process produces a dataset of reliable state--action--point cloud sequences, which subsequently serves as the foundation for training diffusion-based policies.

\subsection{Diffusion Policy with Unsupervised 3D Representation (DP-U3R)}
Additionally, we train our proposed DP-U3R. Let the (state, point cloud, action) sequence collected from RL be denoted as $\tau_s =\{(s_t, p_t,a_t)\}_{t=0}^T$. Each point cloud $p_t$ is perturbed with Gaussian noise $\xi_t \sim \mathcal{N}(0, \sigma^2 I)$ to encourage robust feature learning, yielding $\tilde{p}_t = p_t + \xi_t$. The perturbed point cloud is encoded by an encoder $E(\cdot)$, producing per-point features $f_{i,t} \in \mathbb{R}^d$. A global feature vector $f^g_t \in \mathbb{R}^d$ is then obtained by a pooling operator $P(\cdot)$:
{\small
\begin{equation}
f_{i,t} = E(\tilde{p}_t); f^g_t = P(\{f_{i,t}\}_{i=1}^{M_p}) ,
\end{equation}
}
where $M_p$ is the number of points in the cloud. 
An attention module $A(\cdot)$ fuses local features with the global representation, yielding the final embedding $z_t \in \mathbb{R}^d$:
{\small
\begin{equation}
z_t = A(\{f_{i,t}\}_{i=1}^{M_p}, f^g_t) .
\end{equation}
}
The embedding feature $z_t$ will be used to predict and reconstruct the point cloud $\hat{p}_t$. The diffusion policy receives the augmented state $\tilde{s}_t = [s_t, f^g_t]$ along with the action sequence to model the denoising process.  DP-U3R is optimized by two complementary objectives. The first is the point cloud reconstruction loss, defined as the L2 distance between the original point cloud $p_t$ and reconstruction $\hat{p}_t$:
{\small
\begin{equation}
\mathcal{L}_{\text{recon}} = \frac{1}{M_p} \sum_{i=1}^{M_p} \| p_{i,t} - \hat{p}_{i,t} \|_2^2 .
\end{equation}
}
The second is the noise prediction loss for the diffusion model. 
At each step $t$, the policy predicts Gaussian noise $\hat{\epsilon}_t$ from the noisy action $\tilde{a}_t$, while the ground truth noise is $\epsilon_t \sim \mathcal{N}(0, I)$. 
The loss is given by:
{\small
\begin{equation}
\mathcal{L}_{\text{noise}} = \mathbb{E}_{t, \epsilon_t \sim \mathcal{N}(0, I)} \big[ \|\epsilon_t - \hat{\epsilon}_t\|_2^2 \big] .
\end{equation}
}

The overall training objective is a weighted sum of the two terms:
{\small
\begin{equation}
\mathcal{L} = \lambda_{\text{recon}} \, \mathcal{L}_{\text{recon}} + \lambda_{\text{noise}} \, \mathcal{L}_{\text{noise}} ,
\end{equation}
}
where $\lambda_{\text{recon}}$ and $\lambda_{\text{noise}}$ control the balance between unsupervised learning and diffusion policy optimization.

\subsection{Dynamics-Aware Adaptive Integration Mechanism (DAIM)}
In the final stage, we integrate teleoperation signals from a real glove interface into the diffusion policy via our proposed DAIM. 
Let the action predicted by the diffusion policy at step $k$ be $\hat{x}_k$, and the reference action derived from glove input $h_t$  be $x_{\text{ref}}$. Since diffusion policies operate as conditional denoisers, injecting $h_t$ at inference only modifies the conditioning and does not alter the underlying diffusion prior; empirically, we observe stable denoising across all settings. The integrated action $\tilde{x}_k$ is obtained by a convex combination of the two signals:
{\small
\begin{equation}
\tilde{x}_k = \hat{x}_k + \alpha(k)\,(x_{\text{ref}} - \hat{x}_k) ,
\end{equation}
}
where $\alpha(k) \in [0,1]$ controls the degree of teleoperation intervention. The temporal schedule of $\alpha(k)$ is defined as:
{\small
\begin{equation}
\alpha(k) = \alpha_{\max} \cdot \left(1 - \cos\left(\frac{\pi k}{2K}\right)\right) ,
\end{equation}
}
where $K$ denotes the denoising horizon. The maximum integration weight $\alpha_{\max}$ is determined adaptively based on object dynamics:
{\small
\begin{equation}
\alpha_{\max} = \text{sigmoid}(u_0 - u) ,
\end{equation}
}
where $u_0$ is the adjustment coefficient and the dynamic factor $u$ is computed as
{\small
\begin{equation}
u = \beta_v \cdot \frac{\|v\|}{v_0} + \beta_\omega \cdot \frac{\|\omega\|}{\omega_0} ,
\end{equation}
}
where $v$ and $\omega$ denote the linear and angular velocities of the object, respectively, and $v_0, \omega_0$ are normalization constants. 
$\beta_v, \beta_\omega$ balance the contribution of translational and rotational dynamics. Intuitively, our DAIM adaptively adjusts the strength of teleoperation guidance according to both the diffusion step $k$ and the current object dynamics $(v, \omega)$. When the object moves rapidly, the system relies more on the diffusion policy for stability; when the motion is slower, the teleoperation input is granted more influence. We choose cosine and sigmoid schedules because they preserve the monotonicity and smoothness required by the diffusion process, ensuring that teleoperation acts as a stable conditional guidance rather than a disruptive perturbation.

\section{Experiments}

\subsection{Experiment settings}
\begin{figure}[h] 
  \centering
  % \vspace{-2em}
  \includegraphics[width=0.99\textwidth]{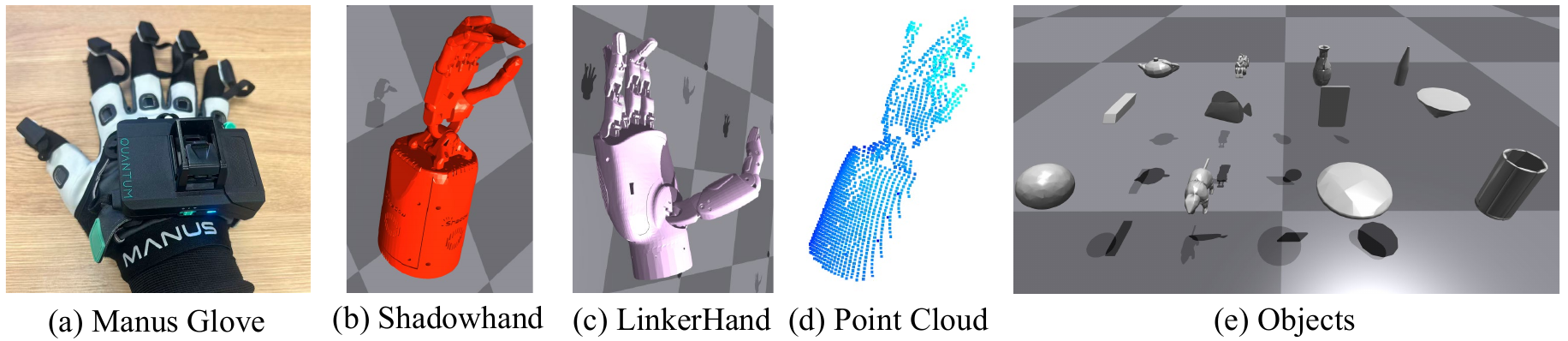} 
  \vspace{-0.5em}
  \caption{Simulation Settings} 
  \vspace{-2em}
  \label{fig:sim_setting}
\end{figure}

\textbf{Simulation Settings.} As illustrated in Fig.~\ref{fig:sim_setting}, our simulation experimental setup integrates both hardware and simulation components. On the hardware side, we employ a Manus glove(Fig.~\ref{fig:sim_setting}(a)), which captures human hand kinematics, including finger joint angles and overall hand pose, providing high-fidelity teleoperation signals. Additionally, all training and inference is conducted on one single NVIDIA RTX 4090 GPU. In simulation, we adopt Isaac Gym as the physics engine, and model the manipulator with a ShadowHand (Fig.~\ref{fig:sim_setting}(b)), a five-finger dexterous robotic hand with 24 degrees of freedom, and a Linkerhand (Fig.~\ref{fig:sim_setting}(c)), a five-finger dexterous robotic hand with 16 degrees of freedom. To provide visual observations, we place an RGB-D camera in the environment, and the captured depth maps are converted into point clouds (Fig.~\ref{fig:sim_setting}(d)), which are uniformly downsampled to 1,300 points per frame. The experimental objects (Fig.~\ref{fig:sim_setting}(e)) are drawn from the public DexGraspNet dataset~\cite{wang2023dexgraspnet}.

\begin{figure}[h] 
  \centering
  \vspace{-2em}
  \includegraphics[width=0.99\textwidth]{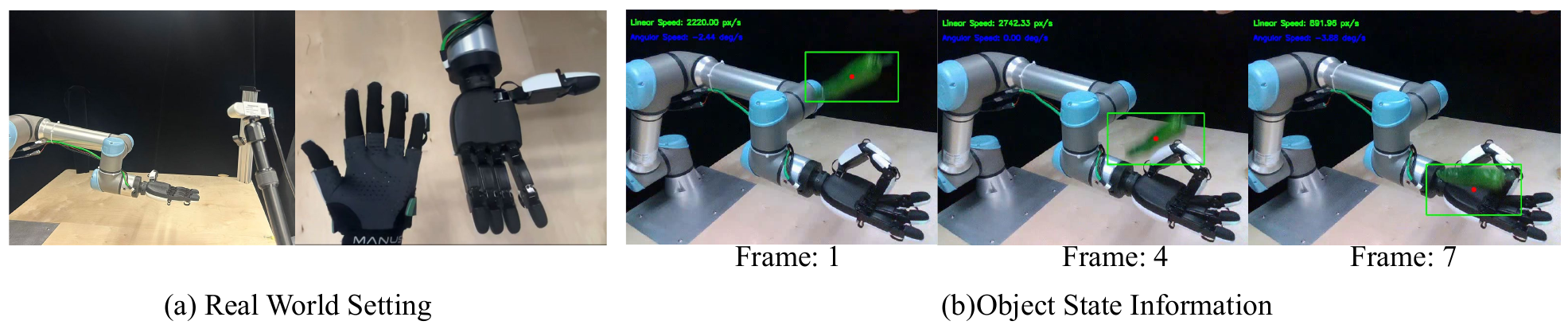} 
  \vspace{-0.5em}
  \caption{Sim2Real Settings} 
  \vspace{-2em}
  \label{fig:sim2real_setting}
\end{figure}

 \noindent \textbf{Sim2Real Settings.} As shown in Fig.~\ref{fig:sim2real_setting}(a), we set up a real-world validation scenario (Xhand left hand + Manus glove + RealSense 435 RGB-D + UR arm). We pre-trained our policy using the Xhand URDF and then deployed it for real-world testing. It is an important and promising direction to explore obtaining the accurate object state in the real world. We recommend using 2D/3D tracking/pose estimation methods to measure object state. In our real-world system, we use the lightweight TrackerCSRT for object tracking and state estimation. As depicted in Fig.~\ref{fig:sim2real_setting}(b), the estimated values for sampled frames are recorded in the top-left corner. 

\noindent \textbf{Implementation Details.} During RL data collection, we construct training and validation sets with a ratio of 5:1. Specifically, for each object category, we sample 5,000 successful trajectories as training data and 1,000 successful trajectories as validation data. For point cloud processing, the encoder outputs per-point features, and the global feature dimension is set to 256. The hyperparameters for DAIM are fixed as $\beta_v = 10$, $\beta_\omega = 0.1$, $u_0=1.0$, $v_0=1.0$, and $w_0=10$. In our DP-U3R training, the loss weights are set as $\lambda_{\text{recon}} = 1.0$ and $\lambda_{\text{noise}} = 1.0$. The detailed RL reward coefficients and their formulations are reported in Appendix.~\ref{sec:reward}. Similar to DP/DP3, our inference time is  41.4 ms with 10 denoise steps on a single RTX 4090. And the glove costs 9.2 ms. The video latency comes from simulation and inference on the same GPU, with the simulation dt setting. The detailed setting of the PPO and the Isaac Gym is recorded in the Tab.~\ref{tab:ppo_hyper} and Tab.~\ref{tab:sim_params} in the Appendix.

\noindent  \textbf{Evaluation Metrics.} Following prior work~\cite{landexcatch,Ze2024DP3,yin2025dexteritygen} in dexterous manipulation, we adopt Success Rate~(\%) as the primary evaluation metric, where a trial is considered successful if the robotic hand catches the object and maintains stable contact for a fixed duration.  We provide the operator with a 15-minute familiarization period before the evaluation. For teleoperation, each object is tested 15 times, and the average success rate is reported.  Moreover, we evaluate the denoising accuracy of the diffusion policy via the Mean Squared Error (MSE) between predicted denoised actions and ground-truth actions, which directly reflects the precision of action prediction.

\subsection{Experiment Results}

\textit{Our goal is to improve teleoperation via shared autonomy, not to optimize the catching task itself.} We report our results in Tab.~\ref{tab:rate}. Pure Teleoperation (Tele-Pure) with glove retargeting attains only 35.3\% due to timing, force, and mapping errors, whereas our Tele-Catch improves performance to 54.7\%, with notable gains on dynamic objects such as Truck and Teapot. Overall, these results validate our dynamics-aware integration while highlighting that human-in-the-loop dynamic catching remains challenging and merits further study.

 \begin{table}[h]
\centering
\vspace{-1.5em}
\resizebox{0.99\textwidth}{!}{
\begin{tabular}{lccccccccccc}
\toprule
Methods & Ball & Bear & Cookie & Cup & Fish & Ipod & Showerhead & Teapot & Truck & Vase & Mean$\uparrow$ \\
\hline
Tele-Pure       & 20.0 & 46.7 & \textbf{80.0} & 26.7 & 6.7 & \textbf{13.3} & 60.0 & 33.3 & 46.7 & 20.0 & 35.3 \\
Tele-Catch  & \textbf{46.7} & \textbf{53.3} & \textbf{80.0} & \textbf{46.7} & \textbf{20.0} & \textbf{13.3} & \textbf{80.0} & \textbf{86.7} & \textbf{73.3} & \textbf{46.7} & \textbf{54.7} \\
\bottomrule
\end{tabular}
}
\caption{Success rate (SR) comparison on all classes. Tele-Pure abbreviates Pure Teleoperation. }
\vspace{-3.5em}
\label{tab:rate}
\end{table}

We find that many failures can be traced back to the four challenges summarized in Fig.~\ref{fig:banner}, including timing, pose alignment, contact force control, and retargeting errors; in particular, thin objects with sharp edges (e.g., Ipod and Fish, Fig.~3) remain challenging: their corners often contact first during falling, causing high-impulse rebounds before a stable enclosure can be formed, which leads to frequent drop failures.

\begin{table}[ht]
\centering
\vspace{-1.5em}
\resizebox{0.9555\textwidth}{!}{
\begin{tabular}{lccccccccccc}
\toprule
& Ball & Bear & Cookie & Cup & Fish & Ipod & Showerhead & Teapot & Truck & Vase \\
\hline
DP      & 0.152 & 0.145 & 0.158 & 0.139 & \textbf{0.136} & 0.155 & 0.156 & 0.156 & 0.161 & 0.140  \\
DP3     & \textbf{0.150} & 0.143 & 0.157 & 0.133 & \textbf{0.136} & 0.155 & 0.159 & \textbf{0.152} & 0.159 & 0.136  \\
DP-U3R  & \textbf{0.150} & \textbf{0.142} & \textbf{0.155} & \textbf{0.129} & \textbf{0.136} & \textbf{0.154} & \textbf{0.154} & \textbf{0.152} & \textbf{0.158} & \textbf{0.134} \\
\bottomrule
\end{tabular}
}
\caption{Action noise MSE results across different objects.}
\vspace{-3.0em}
\label{tab:mse}
\end{table}

 \begin{table}[ht]
\centering
\vspace{-2.0em}
\resizebox{0.9775\textwidth}{!}{
\begin{tabular}{lcc|ccccccccc}
\toprule
  & Open-Loop & RL &Tele-Pure  &  Tele-Catch(DP~\cite{chi2023diffusion})$^{\#}$ &  Tele-Catch(DP3~\cite{Ze2024DP3}) &  Tele-Catch(DP-U3R) \\
\hline
SR $\uparrow$      & 46. 7 &  33.3 & 46.7 & 60.0 & 66.7& \textbf{73.3} \\
\bottomrule
\end{tabular}
}
\caption{Success rate (SR) comparison on Truck. $^{\#}$ stands for our baseline.}
\vspace{-3.0em}
\label{tab:method_rate}
\end{table}

Tab.~\ref{tab:mse} reports the action-noise MSE for each object. Overall, DP-U3R shows lower errors on most categories and smaller fluctuations across objects than the standard diffusion policy and DP3, indicating more reliable denoising and stronger robustness to diverse geometries and dynamics. Since DP3 augments DP with additional 3D cues and DP-U3R further integrates unsupervised point-cloud representations, the results suggest that richer 3D information improves action denoising quality.

This improvement also transfers to task performance. In Tab.~\ref{tab:method_rate}, Pure Teleoperation achieves 46.7\% SR, matching the open-loop DexGraspNet-based baseline~\cite{wang2023dexgraspnet}. With shared autonomy, Tele-Catch improves SR to 60.0\%, 66.7\%, and 73.3\% when using DP, DP3, and DP-U3R, respectively, where Tele-Catch(DP) serves as our ablation baseline for backbone comparison. DexterityGen~\cite{yin2025dexteritygen} does not release its training or evaluation code, so we exclude it from quantitative comparisons under our setting. These results imply that Tele-Catch stabilizes noisy teleoperation commands and reduces timing errors, and that stronger diffusion backbones further improve robustness.

\begin{figure}[ht] 
  \centering
  \vspace{-1.5em}
  \includegraphics[width=0.9775\textwidth]{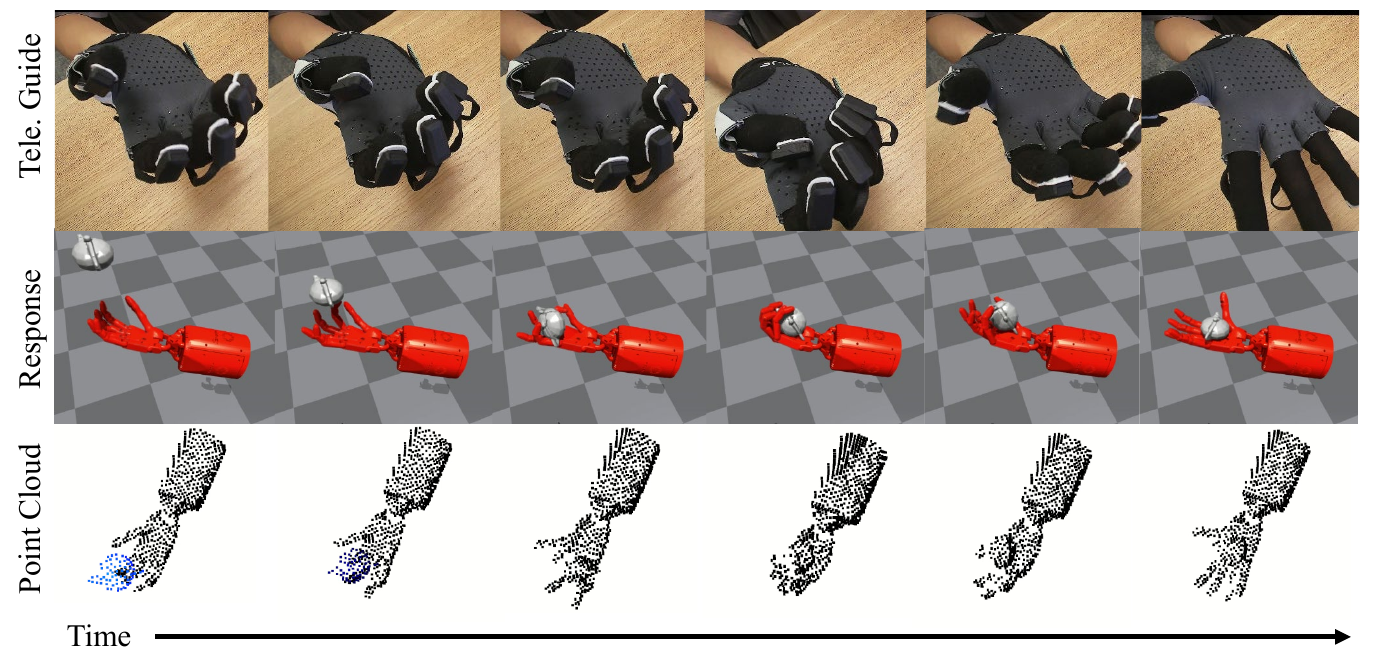} 
  \vspace{-0.5em}
  \caption{Qualitative analysis. The top row shows the teleoperation glove guidance, the middle row depicts the corresponding robotic hand responses in simulation, and the bottom row illustrates the synchronized point-cloud observations. Together, these visualizations demonstrate how Tele-Catch integrates human teleoperation with geometry-aware policy control to achieve stable dynamic object catching over time.} 
  \vspace{-2.5em}
  \label{fig:vis}
\end{figure}

We provide the qualitative analysis of Tele-Catch in Fig.\ref{fig:vis}, showing how glove inputs, robotic responses, and point cloud observations evolve during dynamic catching.  Our method benefits from DP-U3R priors, which provide stable guidance during unstable object motion and ensure more reliable catching. Supplementary video further illustrates this contrast, highlighting the failures of direct teleoperation and the stable performance of Tele-Catch. The supplementary video also demonstrates that our framework can effectively mitigate the four challenges identified in the Section.~\ref{sec:intro}.

\subsection{Sim2Real Analysis}
 Real-world validation is important, and we acknowledge that teleoperation with dynamic objects is still in early stages. Following prior work~[\textcolor{blue}{34}][\textcolor{blue}{3}][\textcolor{blue}{28}],  our experiments are mainly conducted in simulation.  However, our method is designed to be transferable to real-world settings. Here we make an attempt to validate our method in the real-world system.
 
 \begin{figure}[h] 
  \centering
  \vspace{-1.5em}
  \includegraphics[width=0.99\textwidth]{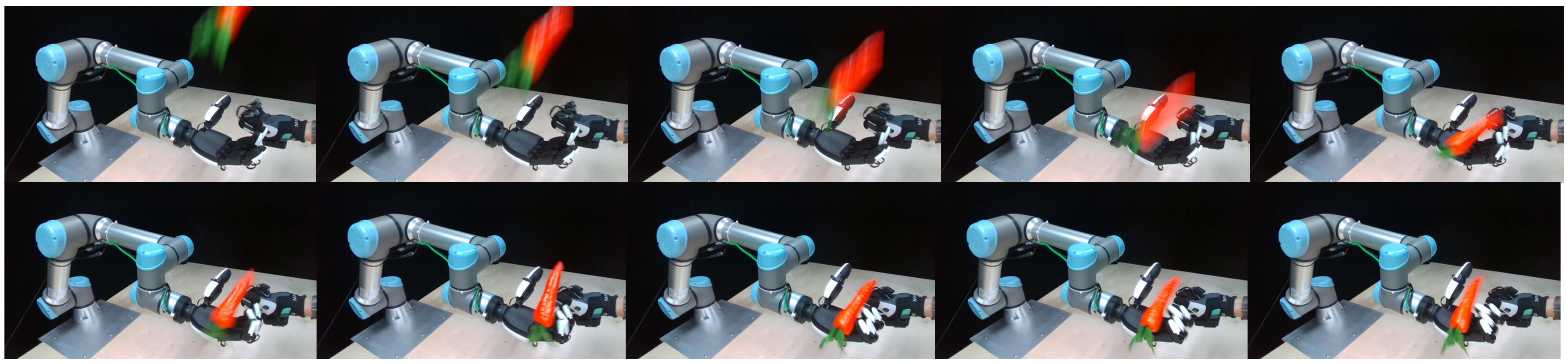} 
  \vspace{-0.5em}
  \caption{Sim2Real analysis} 
  \vspace{-2.0em}
  \label{fig:sim2real}
\end{figure}
 
\begin{figure}[h] 
  \centering
  \vspace{-1.5em}
  \includegraphics[width=0.99\textwidth]{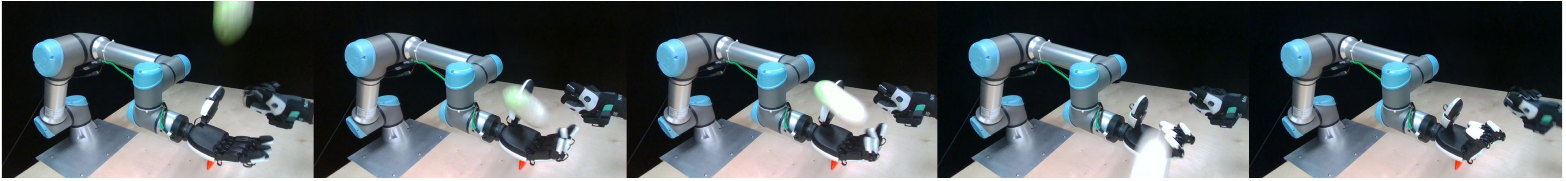} 
  \vspace{-0.5em}
  \caption{Failure case} 
  \vspace{-2.0em}
  \label{fig:fc}
\end{figure}

 We provide a complete, unedited test clip recorded by the RealSense camera, which clearly shows the object, the dexterous hand, and the teleoperation glove throughout the interaction (Fig.~\ref{fig:sim2real}).  Real-world teleoperation exhibits the same challenges described in our paper, often more severely due to light, sensing noise, and contact variability (Fig.~\ref{fig:fc}); thus we prioritize refining and stress-testing in simulation before scaling up hardware evaluations.

\subsection{Generalization Analysis}
Beyond evaluating Tele-Catch under its standard training and execution setting, we further study two complementary aspects of generalization (Tab.~\ref{tab:gen}): (i) cross-embodiment transfer, where Tele-Catch is deployed on a new dexterous hand with different kinematics with retraining, and (ii) unseen-category generalization, where the method is tested on novel object categories without retraining.

\begin{table}[ht]
\centering
\vspace{-1.5em}
\resizebox{0.8225\textwidth}{!}{
\begin{tabular}{lcccc cc}
\toprule
 Objects & ShadowHand & LinkerHand-L20 &Seen & Unseen &  Tele-Pure & Tele-Catch \\
\hline
Teapot    & \checkmark & & \checkmark     &\multicolumn{1}{c|}{}         & 33.3 & \textbf{86.7} \\
Teapot    & &\checkmark &\checkmark      &\multicolumn{1}{c|}{}         & 40.0 & \textbf{73.3} \\ \hline
Eraser    &\checkmark & &     &\multicolumn{1}{c|}{\checkmark}         & 53.3 & \textbf{66.7} \\ 
Telephone &\checkmark& &     &\multicolumn{1}{c|}{\checkmark}         &46.7  &  \textbf{53.3}\\
Bottle    &\checkmark & &    &\multicolumn{1}{c|}{\checkmark}         &  20.0&  \textbf{33.3}\\

\bottomrule
\end{tabular}
}
\caption{Generalization analysis. }
\vspace{-3em}
\label{tab:gen}
\end{table}

\textbf{Cross-embodiment transfer.} 
We evaluate the cross-embodiment performance by transferring Tele-Catch from ShadowHand to LinkerHand-L20. 
Since the kinematics differ, this setting requires retraining the policy on the target embodiment.
Tele-Catch consistently improves success rates compared with pure teleoperation (73.3\% vs. 40.0\% on Teapot), indicating effective cross-embodiment transfer with retraining. 
This highlights the flexibility of Tele-Catch for adapting to new robotic hardware once retrained.

\textbf{Unseen category generalization.} 
We further evaluate Tele-Catch on unseen object categories without retraining, including Eraser, Telephone, and Bottle.
The method consistently outperforms direct teleoperation (e.g., 66.7\% vs. 53.3\% on Eraser), demonstrating non-trivial zero-shot generalization to novel categories.
These results suggest that increasing category coverage during training may further improve performance on everyday objects.

\textbf{Object motion diversity.} 
We clarify that our motion diversity comes from RL data collection and goes beyond pure free-fall; 
for example, adding an initial x-axis velocity yields a 66.7\% success rate on Teapot.

\subsection{Sensitivity Analysis}
\noindent We assign a smaller weight to $\beta_{\omega}$ because the object angular velocity tends to change abruptly during collisions and is generally harder to estimate reliably; down-weighting it makes the controller less sensitive to noisy angular-velocity estimates. On Teapot, sweeping $\beta_v \in \{1,10,20\}$ yields success rates of $60.0/86.7/86.7$, while sweeping $\beta_{\omega} \in \{0.05,0.1,0.15\}$ yields $66.7/86.7/86.7$. When $\beta_v$ or $\beta_{\omega}$ is set too large, the teleoperation command is overly attenuated, leading to conservative responses and degraded catching performance. In our formulation, the blending coefficient $\alpha$ is determined by these $\beta$ weights---larger $\beta$ results in stronger attenuation (smaller $\alpha$), whereas smaller $\beta$ preserves more of the user command.

\begin{figure}[ht] 
  \centering
  \vspace{-1.5em}
  \includegraphics[width=0.9775\textwidth]{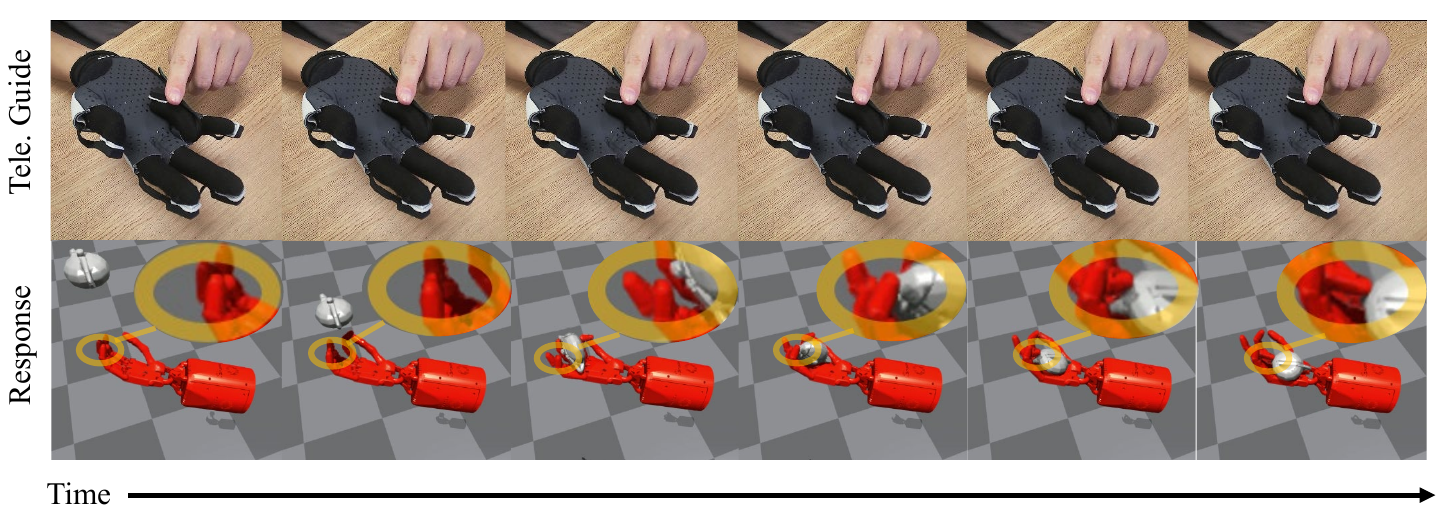}
  \caption{Sensitivity analysis. The orange ellipse highlights the variation of the ring finger.} 
  \vspace{-3.0em}
  \label{fig:sens}
\end{figure}

As illustrated in Fig.~\ref{fig:sens}, we conduct a sensitivity analysis focusing on the ring finger position under teleoperation guidance. We press the ring finger to observe the corresponding deformation of the dexterous hand's ring finger. During the unstable phase of object motion, the robotic hand mainly relies on the diffusion policy to stabilize its response, which enables reliable catching despite rapid dynamics. Once the object is successfully secured and enters a stable state, the teleoperation input gradually becomes dominant, leading the ring finger to bend downward according to the glove command. This behavior highlights the adaptive nature of DAIM, where human guidance and the learned policy are dynamically weighted to ensure stability in catching and responsiveness in teleoperation.

\section{Conclusion}
In this work, we introduced Tele-Catch, the systematic study of dexterous hand teleoperation for dynamic object catching within a shared-autonomy framework. At its core, Tele-Catch leverages DAIM, a dynamics-aware adaptive integration mechanism that fuses glove-based teleoperation inputs into the diffusion policy’s denoising process, ensuring smooth and adaptive control under varying object dynamics. To further enhance robustness and generalization, we proposed DP-U3R, which augments diffusion policy learning with unsupervised 3D point cloud representations, enabling geometry-aware decision making. Experiments demonstrate that Tele-Catch significantly improves success rates over baselines, achieving robust performance across diverse objects and conditions. Moreover, the analysis highlights both the effectiveness of our contributions and the remaining challenges of this new task, pointing to substantial opportunities for future research.

\clearpage

\section{Appendix}

\subsection{Discussion}
\noindent \textbf{Why Teleoperation?} Teleoperation remains a widely adopted and practically valuable paradigm in robotics, enabling human intent to guide manipulation in complex and uncertain scenarios. Our work specifically focuses on improving dynamic object interaction under teleoperation, rather than pursuing fully autonomous solutions. Purely automated dynamic catching cannot be directly substituted into teleoperation pipelines due to inherent retargeting discrepancies, embodiment mismatch, and the difficulty of tightly aligning autonomous actions with human control intent. In contrast, Tele-Catch is designed to enhance teleoperation: it leverages automation not to replace the human, but to compensate for timing, pose, and force errors that frequently occur during human–robot dynamic interaction. Thus, our framework adopts a shared-automation formulation in which autonomous policy priors strengthen stability while teleoperation retains high-level control authority.

\subsection{RL Reward Design Details}
\label{sec:reward}
In this section, we provide the detailed definitions of all reward and penalty terms used in PPO training. The corresponding coefficients are listed in Tab.~\ref{app:reward-coeff}. The fingertip–object distance reward encourages the robotic hand to reduce the spatial gap between its fingertips and the target, thereby accelerating the convergence toward feasible grasp configurations. This guidance not only expedites the catching process but also improves grasp reliability by ensuring timely contact with the moving object. In addition, the fingertip velocity penalties regulate the linear and angular speeds of the fingers, discouraging excessively aggressive motions. By enforcing smoother trajectories, these penalties mitigate overshooting and reduce slippage risks, thereby promoting stable and physically plausible catches. Our reward design enables PPO to learn policies that achieve high catching success but also maintain efficiency and physical plausibility.

\begin{table*}[h]
\centering
\vspace{-1.5em}
\resizebox{0.8225\textwidth}{!}{
\begin{tabular}{llll}
\toprule
\textbf{Component} & \textbf{Symbol} & \textbf{Coefficient} & \textbf{Coefficient Value} \\
\midrule
Hand--object distance reward & $R_{\text{dist}}$ & $\lambda_{\text{dist}}$ & 10.0 \\
Object orientation reward & $R_{\text{rot}}$ & $\lambda_{\text{rot}}$ & 0.1 \\
Fingertip--object distance reward & $R_{\text{ftip-dist}}$ & $\lambda_{\text{ftip-dist}}$ & 10.0 \\
Fingertip linear velocity penalty & $P_{\text{ftip-linvel}}$ & $\lambda_{\text{lin}}$ & 0.3 \\
Fingertip angular velocity penalty & $P_{\text{ftip-angvel}}$ & $\lambda_{\text{ang}}$ & 0.03 \\
Action penalty & $P_{\text{action}}$ & $\lambda_{\text{act}}$ & $2e^{-4}$ \\
Torque penalty & $P_{\text{torque}}$ & $\lambda_{\text{torque}}$ & $2e^{-4}$ \\
Power penalty & $P_{\text{power}}$ & $\lambda_{\text{power}}$ & $2e^{-4}$ \\
Drop penalty (one-time) & $P_{\text{drop}}$ & $\lambda_{\text{drop}}$ & 1.0 \\
\bottomrule
\end{tabular}
}
\caption{Reward components and coefficients used in our RL training.}
\vspace{-3em}
\label{app:reward-coeff}
\end{table*}

\subsubsection{Reward and Penalty Formulations}

Let $p_{\text{hand}} \in \mathbb{R}^3$ be the palm position, 
$p_{\text{obj}} \in \mathbb{R}^3$ the object position, 
$q_{\text{obj}}$ the object orientation quaternion, 
$q_{\text{target}}$ the desired target orientation,
$p_{\text{tip},i}$ the position of the $i$-th fingertip, 
$v_{\text{tip},i}$ its linear velocity, 
$\omega_{\text{tip},i}$ its angular velocity,
$a_t$ the control action at time $t$,
$\tau_t$ the applied torque vector,
and $\dot{\theta}_{t,j}$ the angular velocity of the $j$-th joint.

\begin{itemize}
    \item \textbf{Hand--object distance reward:}  
    \[
    R_{\text{dist}} = - \| p_{\text{hand}} - p_{\text{obj}} \|_2
    \]

    \item \textbf{Object orientation reward:}  
    \[
    R_{\text{rot}} = - \| q_{\text{obj}} - q_{\text{target}} \|
    \]

    \item \textbf{Fingertip--object distance reward:}  
    For $M$ fingertips,
    \[
    R_{\text{ftip-dist}} = - \frac{1}{M} \sum_{i=1}^{M} \| p_{\text{tip}, i} - p_{\text{obj}} \|_2
    \]

    \item \textbf{Fingertip linear velocity penalty:}  
    \[
    P_{\text{ftip-linvel}} = \frac{1}{M} \sum_{i=1}^{M} \| v_{\text{tip}, i} \|_2
    \]

    \item \textbf{Fingertip angular velocity penalty:}  
    \[
    P_{\text{ftip-angvel}} = \frac{1}{M} \sum_{i=1}^{M} \| \omega_{\text{tip}, i} \|_2
    \]

    \item \textbf{Action penalty:}  
    \[
    P_{\text{action}} = \| a_t \|_2^2
    \]

    \item \textbf{Torque penalty:}  
    \[
    P_{\text{torque}} = \| \tau_t \|_2^2
    \]

    \item \textbf{Work penalty:}  
    \[
    P_{\text{work}} = \sum_{j} |\tau_{t,j} \cdot \dot{\theta}_{t,j}|
    \]

    \item \textbf{Drop penalty:}  
    \[
    P_{\text{drop}} = 100
    \]
\end{itemize}

\subsubsection{Reward and Penalty Records}
\begin{figure*}[h] 
  \centering
  \vspace{-1.5em}
  \includegraphics[width=0.95\textwidth]{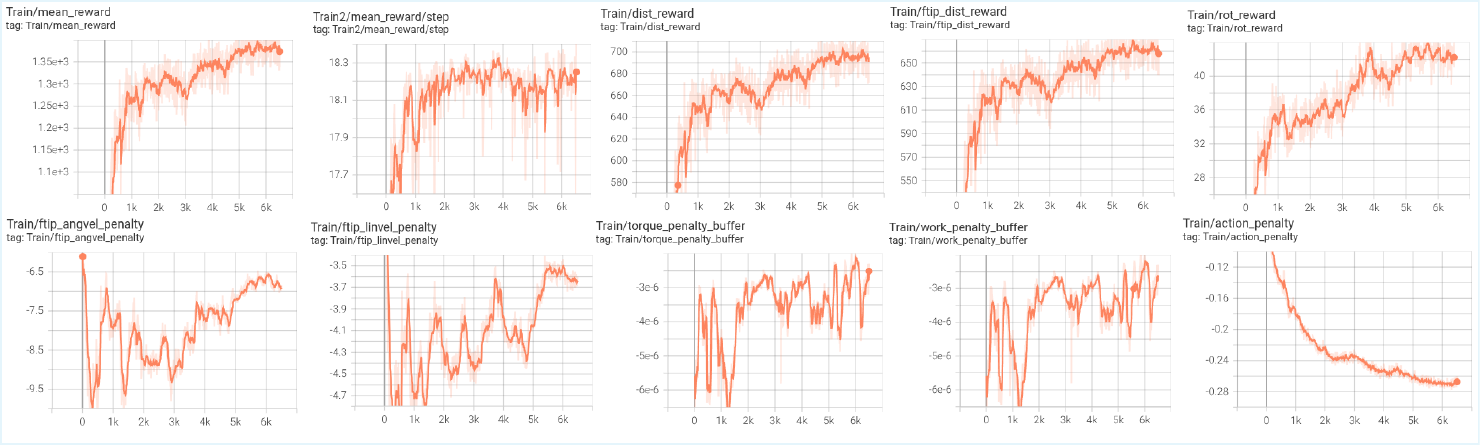}
  \vspace{-1.0em}
  \caption{PPO Reward and Penalty} 
  \vspace{-2.0em}
  \label{fig:rewards}
\end{figure*}

We provide the recorded reward and penalty curves during training, as illustrated in Fig.~\ref{fig:rewards}. The first row shows the major reward terms, including distance reward, rotation reward, and fingertip distance reward. These curves exhibit a clear upward trend and eventually stabilize, indicating that the policy successfully learns consistent catching strategies. The second row corresponds to the auxiliary penalties, such as fingertip linear and angular velocity penalties, torque and work penalties, and action penalties. Since these terms are primarily used to regularize energy consumption and finger motion smoothness, they exhibit higher variability during training. Nevertheless, their overall trends remain bounded, suggesting that the learned policy maintains stable control while avoiding excessive actuation effort.

\subsubsection{PPO hyperparameter setting}

For reproducibility, we summarize in Tab.~\ref{tab:ppo_hyper} the hyperparameter configurations used in training our PPO-based RL baseline. The table includes general settings, policy network architectures, training schedules, optimization details, regularization terms, RL parameters, and logging options. These choices follow standard PPO practices with minor adjustments to ensure stable training in dynamic catching tasks.

\begin{table}[ht]
\centering
\vspace{-1.5em}
\renewcommand{\arraystretch}{0.8} % 压缩行间距
\setlength{\tabcolsep}{6pt}       % 调整列间距
\resizebox{\linewidth}{!}{%
\begin{tabular}{ll}
\toprule
\textbf{Category} & \textbf{Hyperparameters} \\
\midrule
General & Random seed = 22,\; Observation clipping = 5.0,\; Action clipping = 1.0 \\
Policy Network & Actor hidden layers = [1024, 1024, 512] \\
              & Critic hidden layers = [1024, 1024, 512],  \; Activation = ELU \\
Training Setup & Maximum iterations = 6500,\; Schedule = adaptive,\; Save interval = 1000 \\
Optimization & Rollout steps = 8,\; Epochs per update = 5,\; Minibatches per epoch = 4 \\
             & Max gradient norm = 1,\; Learning rate = $3 \times 10^{-4}$ \\
Regularization & Clipping range = 0.2,\; Entropy coefficient = 0,\; Target KL divergence = 0.016 \\
RL Parameters & Discount factor $\gamma = 0.96$,\; GAE parameter $\lambda = 0.95$,\; Initial noise std = 0.8 \\
Logging & Log interval = 1,\; Console logging = True \\
\bottomrule
\end{tabular}%
}
\caption{PPO hyperparameter setting.}
\vspace{-4.0em}
\label{tab:ppo_hyper}
\end{table}

\subsection{Simulation}
\subsubsection{Simulation Setting}
For completeness, we list in Tab.~\ref{tab:sim_params} the key simulation parameters used in Isaac Gym during RL and evaluation. The table summarizes environment configurations, control settings, initialization noise, success criteria, randomization ranges, and physics engine details. These parameters ensure that the training environment is both diverse and stable, facilitating the learning of robust catching strategies.  Such randomized yet bounded settings improve generalization by preventing the policy from overfitting to a single simulation condition.

\begin{table}[ht]
\centering
\vspace{-1.5em}
\renewcommand{\arraystretch}{0.8} 
\setlength{\tabcolsep}{6pt}       
\resizebox{\linewidth}{!}{%
\begin{tabular}{ll}
\toprule
\textbf{Category} & \textbf{Key Parameters} \\
\midrule
Environment & numEnvs = 8192,\; envSpacing = 0.75,\; episodeLength = 50 \\
Control     & dofSpeedScale = 20,\; controlFrequency = 60 Hz \\
Noise (init/reset) & posNoise = 0.01,\; dofPosRandom = 0.2 \\
Task Success & successTolerance = 0.1 \\
Randomization &  hand/object mass [0.5, 1.5],\; friction [0.7, 1.3],\; object scale [0.95, 1.05] \\
Physics (PhysX) & solver = TGS,\; numPosIter = 8,\; contactOffset = 0.002,\; bounceThreshold = 0.2 \\
\bottomrule
\end{tabular}%
}
\caption{Isaac Gym simulation parameter setting.}
\vspace{-4.0em}
\label{tab:sim_params}
\end{table}

\subsection{Pseudo-code of Proposed Method}
To provide a clear overview of our method, we present pseudo-code descriptions of the Tele-Catch pipeline. The framework consists of two complementary phases: a training phase (Alg.~\ref{alg:training}), where RL trajectories are collected and used to train our DP-U3R model, and an execution phase(Alg.~\ref{alg:teleoperation}), where the learned policy is integrated with teleoperation signals through DAIM. The pseudocode aims to clarify the data flow and algorithmic steps beyond the conceptual diagrams given in the main text. 

\textbf{Training Stage.} We employ PPO to generate successful dynamic catching trajectories. These trajectories include state information, executed actions, and synchronized point cloud observations. The collected data are then used to train DP-U3R, which encodes noisy point clouds into geometric embeddings and optimizes a diffusion policy with both reconstruction and noise prediction losses.

\vspace{-1.5em}
\begin{algorithm}[ht]
\caption{Training Stage: RL and DP-U3R}
\label{alg:training}
\begin{algorithmic}[1]
\Statex \textbf{Inputs:} Simulator $\mathcal{E}$, PPO config, point cloud $p_t$
\Statex \textbf{Outputs:} DP-U3R policy $\pi_\theta$

\Procedure{Stage 1: RL Training}{}
  \State Train PPO in $\mathcal{E}$ to catch dynamic objects
  \State Collect successful trajectories $\{(s_t,a_t,p_t)\}$
\EndProcedure

\Procedure{Stage 2: DP-U3R Training}{}
  \State Encode noisy point cloud $p_t$ into embedding $z_t$
  \State Train diffusion policy $\pi_\theta(s_t,z_t)$
  \State Optimize with reconstruction loss + noise prediction loss
\EndProcedure
\end{algorithmic}
\end{algorithm}
\vspace{-2.0em}

\textbf{Execution Stage.} During execution, the trained DP-U3R is combined with glove-based teleoperation inputs using the dynamics-aware adaptive integration mechanism (DAIM). At each control step, the policy proposes an action from the encoded point cloud and state, while glove commands are retargeted to reference actions. DAIM dynamically fuses these two signals according to object dynamics and time step, enabling stable yet responsive teleoperation for dynamic catching.

\vspace{-1.5em}
\begin{algorithm*}[ht]
\caption{Execution Stage: Teleoperation with DAIM}
\label{alg:teleoperation}
\begin{algorithmic}[1]
\Statex \textbf{Inputs:} Trained policy $\pi_\theta$, glove input $h_t$, point cloud $p_t$
\Statex \textbf{Outputs:} Executed action $\tilde{x}_t$

\Procedure{Teleoperate}{}
  \For{control step $t=1 \dots T$}
    \State Observe $(s_t,p_t)$ and compute point cloud feature $f_t^g$
    \State Generate policy action $\hat{x}_t \gets \pi_\theta(s_t,f_t^g)$
    \State Retarget glove input $x_{\text{ref}}$ from $h_t$
    \State Compute DAIM weight $\alpha(k)$ based on dynamics and step
    \State Fuse actions: $\tilde{x}_k = \hat{x}_k + \alpha(k)(x_{\text{ref}}-\hat{x}_k)$
    \State Execute action $\tilde{x}_K$ on the hand
  \EndFor
\EndProcedure

\end{algorithmic}
\end{algorithm*}

% \section*{Acknowledgements}
% Please insert your acknowledgments here.

\bibliographystyle{splncs04}
\bibliography{main}
\end{document}